\documentclass{article}

\usepackage{times}
\usepackage{soul}
\usepackage{url}
\usepackage[hidelinks]{hyperref}
\usepackage[utf8]{inputenc}
\usepackage[small]{caption}
\usepackage{graphicx}
\usepackage{amsmath}
\usepackage{amsthm}
\usepackage{booktabs}
\usepackage{algorithm}
\usepackage{algorithmic}
\title{Counting the Number of Solutions to Constraints}

\author{
Jian Zhang$^1$
Cunjing Ge$^2$
Feifei Ma$^1$\\
$^1$Institute of Software, Chinese Academy of Sciences, China\\
$^2$Johannes Kepler University Linz, Linz, Austria\\
zj@ios.ac.cn,
cunjing.ge@jku.at,
maff@ios.ac.cn
}

\begin{document}

\maketitle

\begin{abstract}
Compared with constraint satisfaction problems, counting problems have received less attention.
In this paper, we survey research works on the problems of
counting the number of solutions to constraints.
The constraints may take various forms, including, formulas in the propositional logic, 
linear inequalities over the reals or integers, Boolean combination of linear constraints.
We describe some techniques and tools for solving the counting problems,
as well as some applications (e.g., applications to automated reasoning, program analysis,
formal verification and information security).
\end{abstract}

\section{Introduction}

Constraint Satisfaction Problems (CSPs) have been studied for decades.
In a CSP, there are some variables, each of which may take a value from some domain, and
a collection of constraints over the variables. To solve the CSP,
one has to find a suitable value for each variable, such that all the constraints hold. 
Examples of CSPs include, among others, graph coloring, the N-Queens problem, and
the SAT problem (see the next section for details).
Over the years, a lot of techniques have been proposed to solve various kinds of CSPs.

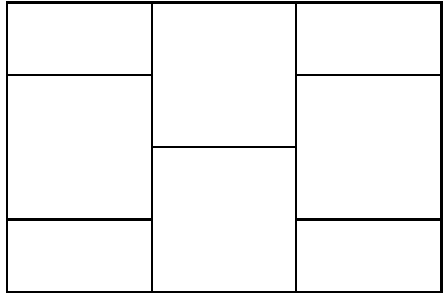
\begin{figure}
\center
\setlength{\unitlength}{1.6mm}
\begin{picture}(42,26)
\put(6,0){\framebox(36,24)}
\put(18,0){\line(0,1){24}}
\put(30,0){\line(0,1){24}}
\put(6,6){\line(1,0){12}}
\put(6,18){\line(1,0){12}}
\put(30,6){\line(1,0){12}}
\put(30,18){\line(1,0){12}}
\put(18,12){\line(1,0){12}}
\end{picture}
\caption{An Exercise for Pupils}
\label{fig:ExPic}
\end{figure}

A different class of problems is the counting version of CSPs (denoted by \#CSP) \cite{CaiLX14}
Instead of deciding the satisfiability of a set of constraints, we would like to count the
number of solutions to the constraints, or to compute the size of the solution space.
Figure~\ref{fig:ExPic} shows a puzzle that has been given to pupils.
There are 8 small rectangles in the big rectangle.
The problem asks: how many different coloring schemes are there, if you can use 4 colors?
Of course, neighboring small rectangles should be assigned different colors.

The counting problems have been studied from various aspects, including their complexity,
algorithms and tools for solving them, and applications in several subareas of AI and computer science.
The related techniques allow us to do more interesting things that are beyond the scope of
traditional constraint solving, like {\em quantitative\/} analysis of systems.

Note that there are different kinds of constraints. A constraint may take the form of logical
formulas, or relations between polynomials, or combinations of logical and arithmetic expressions.
Typically, the theoretical complexity of the counting problem for such constraints can be very high
(\#P-complete or \#P-hard), presenting great challenge for the design of efficient algorithms.
The algorithms 
may be exact or approximate.
In many cases, it is not necessary to obtain the exact counts; a good approximation often suffices.

In this paper, we shall summarize some major research results about counting problems and their applications.
Due to the limit of space, some results (e.g. those related to complexity theory) and references are omitted.
The remainder of this paper is organized as follows.
The next 3 sections present the counting problem and algorithms for the propositional logic,
linear arithmetic constraints and
subsets of first-order logic, respectively.
Section~\ref{sect:AR} and Section~\ref{sect:PA} describe some applications to
automated reasoning and program analysis.
Section~\ref{sect:Misc} covers several other counting problems and applications.
The paper ends with conclusions in Section~\ref{sect:concl}.

\section{\#SAT}\label{sect:sat}

The propositional logic (or the Boolean logic) is a simple formalism;
yet it serves as an important basis for AI and computer science.
In this logic, a formula is constructed from some propositional variables using logical
operators (connectives). A variable or its negation is called a {\it literal}.
A {\it clause\/} is a disjunction of literals, e.g., $(p \vee \neg q)$.
A propositional formula is said to be in the conjunctive normal form (CNF),
if it is a conjunction of clauses.
The satisfiability problem (SAT) can be regarded
as a special kind of CSP, where each variable can be assigned one of the
two truth values, and the constraint is a propositional formula in CNF.

An assignment which makes the formula true is called a {\em satisfying assignment\/} or a {\em model}.
It can be represented by a set of literals. For example, $\{ p, \neg q \}$ means assigning TRUE to $p$
and assigning FALSE to $q$.
Given a propositional formula, the problem of {\em model counting\/}
is to calculate the number of models of the formula. It is often denoted by \#SAT.

\paragraph{Example.} Let us look at the propositional formula (in CNF): $ (p \vee \neg q) \wedge (q \vee \neg p) $. The assignment \{$p$, $q$\} satisfies the formula, so it is a model of the formula.
The formula has two models. The other one is \{$\neg p$, $\neg q$\}, i.e., assigning FALSE to $p$ and to $q$.
So, for this simple formula, the answer to the satisfiability problem is YES (Satisfiable);
while the answer to the model counting problem is 2.

For the propositional logic, the model counting problem is \#P-complete \cite{Valiant79}.
It is widely believed that \#SAT is significantly harder to solve than SAT.
\#SAT is intractable even for Horn and monotone formulae, and even when the size of clauses and number of occurrences of a variable in the formula are extremely limited \cite{Roth96}.

Despite the hardness of the problem, researchers have developed some practical methods for solving \#SAT. They can be put into two categories: exact counting and approximate counting. Most approaches to exact counting try to adapt existing DPLL-style SAT solvers. Strategies like connected component partition and caching were introduced and implemented. A different approach for exact counting is to convert or compile the given formula into a form (e.g., the decomposable negation normal form) from which the model count can be deduced easily.
For approximate counting, local search based or DPLL-based methods with MCMC sampling were proposed. They are categorized into guarantee-less counters. There were also some efforts on bounding counters (estimations with lower bounds or upper bounds) including improvements on the previous guarantee-less counters and XOR hashing-based methods.
For more details, see \cite{GomesSS09} which is a survey of research works (up to the year 2009) on propositional model counting.

After 2009, there were some new works on compilation-based algorithms like \texttt{Dsharp} \cite{MuiseMBH12},
\texttt{SDD} \cite{Darwiche11}
and \texttt{cnf2obdd} \cite{TodaT15}.

\cite{ChakrabortyMV13} extended the hashing-based algorithms from bounding counters to the $(\epsilon, \delta)$-counter which guarantees the estimation lies in interval $[(1+\epsilon)^{-1}\#F, (1+\epsilon)\#F]$ with probability at least $1-\delta$, where $\#F$ is the exact count. Later, more improvements were proposed for this algorithm, 
such as the binary-search strategy \cite{ChakrabortyMV16b}, 
short XOR functions \cite{ZhaoCSE16}, 
and the in-processing XOR-gate recover technique \cite{SoosM19}.
\cite{GeMLZM18} proposed a different framework of hashing-based algorithm, which requires only one satisfiability query for each hash and could terminate once the criterion of accuracy is met. 
\cite{SharmaRSM19} introduced a hashing-based component caching strategy. They implemented a probabilistic exact counter which returns exact count with a least probability, based on the exact counter \texttt{sharpSAT}.

Besides the direct improvements on counting algorithms, \cite{LagniezLM16} investigated the preprocessing of propositional formulas. Experiments showed that such techniques are quite effective for exact model counters.

The classical model counting problem for the propositional logic can be extended in various ways.
\begin{itemize}
\item
Weighted model counting: the propositional variables are assigned some weights, and the problem is to compute a weighted model count of the given formula. We will describe it in more detail, in Section~\ref{sect:AR}.
\item
Maximum model counting (Max\#SAT) \cite{FremontRS17}, which generalizes both MaxSAT and \#SAT.
\item
Projected model counting, \cite{AzizCMS15}, 
i.e., determining the number of models of a propositional formula after eliminating from it a given set of variables.
\end{itemize}

\section{Solution Counting for Linear Constraints}\label{sect:lc}

A linear constraint (LC) is an expression that may be written in the form $a_1 x_1 + a_2 x_2 + \dots + a_n x_n\ op\ a_0$, where $x_i$'s are numeric variables, $a_i$'s are constant coefficients, and $op \in \{<, \le, >, \ge, =, \neq \}$. It is also called a linear arithmetic constraint. As the fundamental constraints that are used in many areas, LCs have been studied thoroughly. The solution counting over LCs is also an important problem.
The numeric variables of LCs can be all reals or all integers. The definition of the counting problem and
the solving techniques are different for different types of variables.

\subsection{Linear Constraints over Reals}

For linear constraints over reals, it is natural to consider the volume of the solution space. Since the solution space of a set of LCs is a convex polytope, the counting problem can also be regarded as the volume computation or estimation problem for polytopes.

\subsubsection{Volume Computation}

As a classical problem in mathematics, there are already several basic approaches about volume computation for convex polytopes.
\cite{MI94-07} discussed several basic approaches in depth.
A polytope can be described in two ways: H-representation (a set of LCs) and V-representation (a set of vertices of the convex hull). The theoretical complexity of the volume compuation is known to be \#P-hard when a polytope is given by only one (either V- or H-) representation \cite{DyerF88}.
But it is unknown for problems given by both.
Moreover, the transformation between the two representations is itself a hard problem which has been studied thoroughly \cite{FukudaLM97}.
Some algorithms require only one and some requires both representations.
In this section, we focus on LCs, i.e., H-representation.

\cite{Bueler2000} observed that previous approaches can only work very efficiently on some polytopes, so they designed a hybrid method to combine advantages from two algorithmic classes. They also implemented a tool called \texttt{Vinci} which embodies five algorithms.
Among them, only Lasserre's method requires only H-representation. It is a signed decomposition method that decomposes a given polytope into signed simplices such that the signed sum of their volumes is the volume of the polytope. This method can handle problems with around 10 dimensions very efficiently. However, it may quickly run out of memory when the number of variables grows slightly.


\subsubsection{Volume Estimation}

Sometimes we only need to {\em estimate\/} the volume of convex polytopes. 
By uniformly generating sample points and counting the number of points that lie in the solution space, the Monte-Carlo method is a straightforward way to estimate the volume of a convex body.
Based on a direct Monte-Carlo method, \cite{LiuZZ07} developed a tool for convex polytopes.
\cite{BorgesFdPV14} introduced an algorithm based on this idea. They implemented a tool called \texttt{qCORAL}. 
It can handle arbitrary constraints, such as non-linear polynomial constraints, trigonometric constraints, etc.
However, it suffers from the curse of dimensionality, 
which means the possibility of sampling inside a certain space in the target object decreases very quickly when the dimension increases.  
As a result, the sample size has to grow exponentially to achieve a reasonable estimation.

To avoid the curse of dimensionality, \cite{DyerFK89} proposed a polynomial time randomized approximation algorithm (FPRAS) called Multiphase Monte-Carlo Algorithm. It first employs a set of concentric balls $B_0 \subset B_1 \subset \dots \subset B_l$ to cut the given convex body $C$ and obtains $B_0 = K_0 \subset K_1 \subset \dots \subset K_l = C$, where $B_0 \subset C \subset B_l$ and $K_i = B_i \cap C$. Note that $vol(C) = vol(K_0) \cdot \prod \frac{vol(K_{i+1})}{vol(K_{i})}$. Then it uniformly generates sample points by a {\em hit-and-run} random walk method, and employs Monte-Carlo method to approximate the ratio $\frac{vol(K_{i+1})}{vol(K_{i})}$. Finally, it multiplies those approximations of ratios with the volume of a ball in $C$ which is easy to compute.

At first, the theoretical complexity is $O^*(n^{23})$\footnote{The ``soft-O'' notation $O^*$ indicates that we suppress factors of $\log n$ as well as factors depending on other parameters.} membership oracle calls. It was reduced to $O^*(n^3)$ at last by a series of works~\cite{CousinsV15}. The membership oracle is a subroutine to detect whether a point is in the convex body. Despite the polynomial time results and reduced complexity, there was still a lack of practical implementation for a long period.

\cite{LovaszD12} is an early experimental implementation of the Multiphase Monte-Carlo algorithm. They reported that the theoretical number of steps for hit-and-run method mixing is much higher than actually required according to the numerical experiences. It indicates that implementing a scalable tool is possible.

\cite{GeM15} focused on convex polytopes. They found that the {\em coordinate directions} hit-and-run method outperforms the commonly used hypersphere directions method with respect to not only runtime, but also accuracy. Besides, they proposed a re-utilization technique for sampling points which was proved no side-effect on the error \cite{GeMZZ18}.
A practical tool called \texttt{PolyVest} was implemented, which can efficiently handle general polytopes with dozens of dimensions.

\cite{CousinsV16} implemented the algorithm proposed in \cite{CousinsV15} whose theoretical complexity is $O^*(n^3)$ oracle calls and $O^*(n^5)$ arithmetic operations. They considered some special convex bodies, such as cubes, half-balls, Birkhoff polytopes, etc, whose oracles are simple. On these convex bodies, the runtime grows roughly like $n^{2.5}$, so that it can handle 100-dimensional cases.

\subsection{Counting Integer Solutions}

The integer solution counting problem for LCs is equivalent to the problem of counting integer points inside convex polytopes. We consider two classes of approaches: exact counting and approximate counting.

\subsubsection{Exact Counting}

\cite{Barvinok93} gave an algorithm that counts integer points in convex polytopes in polynomial time when the dimension is fixed. \cite{LoeraHTY04} described the first implementation of this algorithm, \texttt{LattE}. Like the exact volume computation algorithm on convex polytopes, Barvinok's algorithm is also a signed decomposition method whose running times are very sensitive to the number of dimensions. From experiments, we observe that \texttt{LattE} can deal with instances up to around 16 dimensions and 7 dimensions for hard random cases (smaller than \texttt{Vinci}). \cite{VerdoolaegeSBLB07} presented a new implementation of Barvinok's algroithm, called \texttt{barvinok}. They introduced a parametric version of Barvinok's algorithm for parametric polytopes.
It was proved practically more efficient for problems arising in programming language context.

\subsubsection{Approximate Counting}
\cite{KannanV97} proposed a polynomial time random walk method, which could nearly uniformly generate integer sample points in the polytope. By integrating this into Multiphase Monte-Carlo algorithm, one can obtain a FPRAS algorithm for integer counting. However, this sampling method requires that the given polytope has to contain a ball of radius $n \sqrt{\log m}$, where $m$ is the number of facets and $n$ is the number of variables. For such polytopes, they also showed that the volume approximates the integer counts to within a constant factor.

\cite{GeMMZHZ19} further investigated the problem of approximating the integer counts via the volume. They observed that the execution time of volume computation or estimation is much smaller than integer counting, while output values of volume and integer counts are close in many cases. However, counterexamples show that approximating integer counts via volume with arbitrary precision is impossible. \cite{GeMMZHZ19} proposed and proved a bound for approximation, i.e., the difference between the volume and the integer count. The overhead of computing the bound is negligible.

\section{\#SMT}\label{sect:smt}

Although the propositional logic is an important basis for AI and computer science,
it is inconvenient to use the logic directly for many applications.
Early this century, the Satisfiability Modulo Theories (SMT) paradigm emerged.
It is an extension to SAT.
It checks the satisfiability of logical formulas with respect to combinations of background theories.
Such theories include the theories of linear arithmetic (LA), bit vectors (BV), arrays, strings, and so on.
The solution counting problem for SMT formulas is called \#SMT.

\subsection{\#SMT(LA)}

SMT(LA) formulas are Boolean combinations of linear arithmetic constraints, e.g.,
$(x-y \leq 2) \wedge (x+y = 8) \vee (x>6)$.
An SMT(LA) formula $F$ can be represented as a Boolean formula $PS_F(b_1, \dots, b_n)$ together with definitions in the form: $b_i \equiv c_i$, where $c_i$s are LCs. $PS_F$ is called the propositional skeleton of $F$. For LCs over integers, the SMT(LA) formula is also called an SMT(LIA) formula. Similarly, for LCs over reals, the formula is called an SMT(LRA) formula. And the \#SMT(LRA) problem is to find the volume of the solution space of an SMT(LRA) formula.


\cite{MaLZ09} proposed an algorithm for computing the volume of the solution space of SMT(LRA) formulas.
Since
$$vol(F) = \sum_{\alpha \in Model(PS_F)} vol(\alpha),$$
a straightforward way is to enumerate feasible assignments for $PS_F$ and add the volume of polytopes that correspond to those assignments. \cite{MaLZ09} employed DPLL(T) framework to find the feasible assignments in $Model(PS_F)$ and invoked volume computation tools for convex polytopes to compute $vol(\alpha)$. Moreover, they introduced a new technique called bunch strategy. Each time a feasible assignment $\alpha$ for $PS_F$ is obtained from DPLL engine, it will search the neighborhood of $\alpha$ by negating literals and try to combine $\alpha$ with its feasible neighboring assignments. Then a partial assignment that still propositionally satisfies $PS_F$ is obtained. The resulting assignment may cover a bunch of feasible assignments, hence it is called a ``bunch''. At last, the volume computation subroutine is called for the polytope corresponding to each bunch rather than each feasible assignment, so that the number of calls to polytope volume computation is reduced.

For SMT(LIA) formulas, we also have
$$\#F = \sum_{\alpha \in Model(PS_F)} \#\alpha.$$
Therefore, by integrating volume estimation, exact integer counting and approximate integer counting methods introduced in Section~\ref{sect:lc} with DPLL(T) framework, one can estimate the volume of the solution space of SMT(LRA) formulas, and compute or approximate integer solution counts of SMT(LIA) formulas, respectively. \cite{GeMZZ18} presented such an extension, and implemented a tool called \texttt{VolCE} for \#SMT(LA) problems. Besides, they also introduced several strategies and preprocessing techniques on convex polytopes to improve the performance.


\cite{ZhouHSHCG15} proposed a BDD-based search algorithm for \#SMT(LRA) which reduces the number of conjunctions of LCs. For each conjunction, they introduced a Monte-Carlo integration with a ray-based sampling strategy to approximate the volume. 
The integration algorithm is essentially a direct Monte-Carlo method with exponential time complexity. 
\cite{GaoLZC18} proposed a Markov Chain Monte-Carlo (MCMC) algorithm with the flat histogram method for \#SMT(LIA). The flat histogram method is based on the observation that if a random walk in the energy space is performed with a probability proportional to the reciprocal of the density of states $\frac{1}{g(E)}$, then a flat histogram is achieved for the energy distribution. The energy function $g$ here is a function for any assignment $\alpha$, $g(\alpha) = 0$ if $\alpha$ is a solution of the SMT(LIA) formula and $g(\alpha) > 0$ otherwise. Both methods lack guarantees (bounds) for estimations.
The method lacks guarantees (bounds) for estimations.

\subsection{\#SMT(BV)}

Hashing-based approximate methods for model counting (\#SAT) have been paid much attention in recent years. Due to the deep relation between propositional logic formulas and SMT formulas, it is natural to apply such model counting algorithms to \#SMT.

\cite{ChistikovDM15} thus introduced a bit-level hashing-based approximate counter. Smiliar to the hashing-based approximate counter for \#SAT, it uses XOR-based bit-level hash functions to propositionalize the solution space of an SMT formula and obtain a randomized subset of the solution space. Then it calls an SMT solver repeatedly to count the subset and estimates the total counts with the approximate ratio of the proportion. Even though it is a general framework for SMT formulas with a wide range of theories, it is only practical for bit-vectors due to the limitations of XOR reasoning in state-of-the-art solvers.

\cite{ChakrabortyMMV16} proposed a word-level hashing-based approximate counter for \#SMT(BV), called \texttt{SMTApproxMC}. It views bit-vectors as numbers and generates hash functions with modular arithmetic. So it benefits from the word-level reasoning of state-of-the-art SMT(BV) solvers. \texttt{SMTApproxMC} outperforms the bit-level hashing-based approximate counter \cite{ChistikovDM15}, especially, over benchmarks with more arithmetic constraints.

Recall in Section \ref{sect:sat}, \cite{GeMLZM18} proposed a framework of hashing-based algorithm which is different from previous works. They implemented a tool called \texttt{STAC} not only for propositional logic, but also for SMT(BV) formulas. It outperforms \texttt{SMTApproxMC} by one to two orders of magnitude. The $(\epsilon, \delta)$-style bounds provided by \texttt{STAC} are well supported by experimental results, but lack a theoretical proof.



\section{Probabilistic Inference}\label{sect:AR}


Reasoning with uncertain knowledge and beliefs is an important research area in AI.
There are different kinds of approaches and methodologies in this area,
among which, the probabilistic approach is quite popular.
In particular, a knowledge representation framework called {\it Bayesian networks\/} has attracted much attention.

A Bayesian network (BN) is a directed acyclic graph (DAG), where the nodes represent some random variables, and the arcs represent probabilistic dependencies among the variables.
Associated with each node is a conditional probability of the variable given its parents.
The arcs indicate direct influence of the variables.

For any knowledge representation formalism, we should be concerned with the convenience of
obtaining and expressing knowledge with this formalism, and the efficiency of
using the knowledge to get conclusions (or to answer queries).
For the latter, researchers have proposed quite some inference algorithms (e.g., variable elimination)
for Bayesian networks.

Since exact Bayesian inference is known to be \#P-complete \cite{Roth96}, it is natural to consider
solving the problem via model counting.
\cite{SangBK05,ChaviraD08} described how to transform some probabilistic inference problems to
the weighted model counting (WMC) problem in the propositional logic.
The approach has been shown to be quite powerful in certain cases.

In WMC, we assign some weights to the propositional variables, which induce a weight for each model.
Specifically, for variable $x$, we assign a weight $w(x) \in [0,1]$;
and the literal $\neg x$ got the weight $1-w(x)$.
The weight of a truth assignment is the product of the weights of its literals.
The weighted model count of a formula is the sum of the weights of
satisfying assignments.

\section{Program Analysis}\label{sect:PA}

Given a program, it is important to know what kind of ``good'' properties it has,
what kind of ``bad'' properties it does not possess.
An example of good properties is termination;
and an example of bad properties is null-pointer dereference.
A lot of research works have been done to formally verify that a program has certain properties.
On the other hand, some researchers have proposed various methods for analyzing or testing programs
to identify bugs in them. 

\subsection{Symbolic Execution}

Among the approaches for program analysis and testing,
a powerful technique is {\em symbolic execution}. Its basic idea is to ``execute'' the program
by initializing variables with symbolic (rather than concrete) values.
During the execution, each variable's value is kept as a symbolic expression,
in terms of the initial values of the input variables.
For instance, suppose the program has only one input variable $n$, whose initial value is denoted by $n_0$.
After executing the assignment statements 
\begin{verbatim}
    i = n+1;
    j = 2*i-1; 
\end{verbatim}
the variables $i$ and $j$ will take the values: $i = n_0 +1$,  $j = 2*n_0 +1$, respectively.

A program can be represented by its flow graph which is a directed graph.
Each path in the graph starts with the beginning (entry) of the program. It consists of a sequence of
statements or conditional expressions.
Symbolic execution of a program path will result in a set of constraints, called the {\it path condition},
which describes the constraints on the initial values of the input variables.
Any input data satisfying the path condition will drive the program to be executed along that path.
Thus, symbolic execution of one path may correspond to many real executions (or, testing the program
many times). Of course, if the path condition is not satisfiable, the corresponding path is not executable.


In some sense, a program's behavior is the union of the behavior of all the paths.
If every path works correctly, the program is proved to be correct.
In practice, we can only check a small number of paths, for non-trivial programs.
Even so, quite some powerful tools for program analysis and testing have been developed,
which are based on symbolic execution of program paths.
Constraint solvers (SMT solvers) play a key role in such tools -- the solvers
can be used to check the satisfiability of path conditions.

\subsection{Path Execution Probability}

When a program path is executable, a question arises: how many input data make the
program execute along that path? Or, how often is the path executed? 
(Suppose that the input space is uniformly distributed.)
This information can be obtained by combining symbolic execution with solution counting.
For example, suppose that at a certain point in a program path, the path condition is
$(n \leq 50) \wedge (n > 20)$, and the next statement is a conditional statement: 
\begin{verbatim}
    if (n-10 > 30) {...} else {...}
\end{verbatim}
Here $n$ is an integer variable. It is easy to know that the {\tt if-then} branch is executed
less frequently than the {\tt else} branch. The first branch corresponds to the constraint
$(n \leq 50) \wedge (n > 20) \wedge (n-10 > 30)$, which has 10 solutions; while the constraint
for the second branch has 20 solutions.

Given that path execution probability can be calculated, we can estimate a program's
behavior by a weighted sum of the behavior of its paths. Here each weight is
the execution probability of a path in the program.

\cite{GeldenhuysDV12} presented the so-called {\em probabilistic symbolic execution}.
It was implementated using Symbolic PathFinder (SPF, a symbolic execution tool for Java programs)
and \texttt{LattE} (which counts integer solutions).
The prototype tool is restricted to linear integer arithmetic.
It can be used to estimate the reliability of programs.
The reliability can be calculated as: the total number of solutions to the path conditions
for all the correct paths, divided by the size of the whole input space.

\subsection{Quantitative Information Flow Analysis}

Information leak is a severe problem for IT systems, and it has been an important research topic
for decades.
Still, for complicated systems, it is impractical to prevent information leaks completely. 
In the past 10 years, researchers became interested in
{\em quantitative information flow\/} (QIF) analysis,
which tries to give a measure of leaked information.

\cite{Backes09} presented an automatic method for information-flow analysis that discovers
what information is leaked and computes its comprehensive quantitative interpretation.
The leaked information is characterized by an equivalence relation on secret artifacts,
and is represented by a logical assertion over the corresponding program variables.
A procedure was given, to compute the number of discovered equivalence classes and their sizes.
In the experiments, \texttt{LattE} was used to count the number of solutions to each
conjunction of linear inequalities. 

More recently, \cite{KlebanovMM13} described an approach to QIF for imperative programs,
which combines bounded model checking, \#SAT solving and SAT preprocessing. 
Their tool can be used, for example, to evaluate image anonymization techniques.
\cite{FremontRS17} applied Max\#SAT to solve the problem of finding the largest information leak in a program with an adversary-controlled input.

\section{Other Problems and Applications}\label{sect:Misc}

\subsection{Counting-Based Search Heuristic}

We all know that heuristics are very important in solving constraint satisfaction and
optimization problems. Some researchers (e.g., \cite{PesantQZ12}) advocate {\em counting-based
search}, which employs the number of solutions information as a heuristic.

Traditionally, most search heuristics in constraint solving rely on local information 
such as the domain sizes of variables.
In counting-based search, we may use more heuristics which are based on the number of solutions.
For example, during the search, we may focus on the constraint currently having the smallest number of solutions.
\cite{PesantQZ12} proposed algorithms for counting the number of solutions to specific families of constraints (e.g. {\tt alldifferent}).


\subsection{Counting Linear Extensions}

Suppose we are given a set of elements coupled with a binary relation that defines the mutual order between some elements. An interesting
question is: how many possible ways are there to extend this partially ordered set into a linear order, where all pairs of elements
are comparable? The problem of counting the linear extensions of a given partial order is \#P-complete. It arises in numerous applications, including sorting,
sequence analysis, preference reasoning, partial order plans, and learning graphical models.

Dynamic programming has been successfully applied in counting linear extensions, but it requires exponential time and space in the worst case. Several fully polynomial-time randomized schemes have also been designed, which are based on Markov chain Monte Carlo (MCMC) schemes such as the telescopic product estimator and the Tootsie Pop algorithm. Recently, 
\cite{TalvitieKNK18b} present a novel scheme called relaxation Tootsie Pop, and showed that it
is superior to all previous schemes (including the scheme of encoding the problem into \#SAT).

\subsection{Temporal Planning}

Temporal planning is an important task in AI. Simple temporal networks (STNs) are traditionally used to guide scheduling decisions during execution. 
For an STN, a desirable feature is its flexibility.
It characterizes the room within a schedule for an agent to reschedule without violating the overall time constraints in the case of unexpected opportunities or disruptions. It is crucial in maximizing an agent's autonomy.

An STN with $n$ events can be represented as a polyhedron in $n$-dimensional space. Any valid schedule corresponds to a point inside the polyhedron. The relative volume of the polyhedron captures the size of the STN's solution space, hence can be used to measure the flexibility of the STN\cite{PolicellaSCO04}. \cite{HuangLOB18} further proposed to adopt the ratio of volume to surface area of the polyhedron, which approximates the ratio of the number of solutions to the number of possible schedule failure points, as a new metric for flexibility. The advantage of this metric is that it captures the distribution of solutions.

\subsection{Quantitative Verification of Neural Networks}

The last decade has seen dramatic advances in artificial neural networks.
Such networks are expected to be employed in many applications, some of which are
safety-critical (e.g., self-driving cars).
It is important to verify these neural networks.
However, unlike conventional formal verification which ensures a program works correctly
for all input data, for neural networks, the existence of a few counterexamples seems acceptable.

Instead of full verification, 
\cite{BalutaSSMS19} studied the {\em quantitative\/} verification of neural networks.
More specifically, the verification problem is the following:
Given a set of neural networks $N$ and a property of interest $P$,
determine how often $P$ is satisfied.
The property $P$ can be defined over the union of inputs and outputs of neural networks in $N$.

\cite{BalutaSSMS19} proposed an approach which encodes a {\it binarized\/} neural network into a 
propositional logical formula.
To determine how often $N$ satisfies $P$, the approach generates a set of logical formulas, $\phi$,
from $N$ and $P$, and then solves the model counting problem over $\phi$.
Their prototype tool NPAQ can be used to evaluate a neural network's
robustness, fairness and effectiveness of trojan attacks.

\section{Concluding Remarks}\label{sect:concl}

Counting is a fundamental problem in mathematics, computer science and AI.
It has been studied in vaious forms, by researchers in different areas.

In this paper, we summarized some research results about the counting problem for the propositional logic, 
subsets of first-order logic, linear arithmetic constraints and other specific constraints.
We focus on \#SAT, \#LC and \#SMT, since the corresponding constraints appear frequently.
We also described the applications of counting techniques to probabilistic inference, constraint solving,
quantitative program analysis and verification, and so on.
Some researchers are exploring other applications, e.g., program synthesis \cite{FremontRS17}.
The diversity of the applications has shown great potential for research on counting problems.
We believe that this is an important emerging area, and
the related techniques and tools will help advance the state-of-the-art in various fields.

While significant progress has been made in the past few decades, the major obstacle that prohibits constrained counting from meeting the increasing demands in various applications remains to be the scalability issue. It is highly desirable to push the limits of counting tools even further.
Another issue is the expressiveness of the constraints. Most existing works focus on (near-)propositional logic
or simple integer constraints. It is quite challenging to tackle counting problems for more complex constraints. 
There are also some specific research problems to work on. For example, can we find better approximation schemes with tight theoretical guarantees for (weighted) counting? How shall we perform constrained counting efficiently in an incremental way?

\section{Acknowledgement}
This paper has been submitted to the survey track of IJCAI-20.
We would like to thank the reviewers for their helpful comments.

\bibliographystyle{plain}
\bibliography{ref}

\begin{thebibliography}{10}

\bibitem{AzizCMS15}
R.~Aziz, G.~Chu, C.~J. Muise, and P.~J. Stuckey.
\newblock {\#}{\(\exists\)}sat: Projected model counting.
\newblock In {\em Proc. of {SAT}}, pages 121--137, 2015.

\bibitem{Backes09}
M.~Backes, B.~K\"{o}pf, and A.~Rybalchenko.
\newblock Automatic discovery and quantification of information leaks.
\newblock In {\em Proc. of S\&P}, pages 141--153, 2009.

\bibitem{BalutaSSMS19}
T.~Baluta, S.~Shen, S.~Shinde, K.~S. Meel, and P.~Saxena.
\newblock Quantitative verification of neural networks and its security
  applications.
\newblock In {\em Proc. of {CCS}}, pages 1249--1264, 2019.

\bibitem{Barvinok93}
A.~I. Barvinok.
\newblock A polynomial time algorithm for counting integral points in polyhedra
  when the dimension is fixed.
\newblock In {\em {FOCS}}, pages 566--572, 1993.

\bibitem{BorgesFdPV14}
M.~Borges, A.~Filieri, M.~d'Amorim, C.~S. Pasareanu, and W.~Visser.
\newblock Compositional solution space quantification for probabilistic
  software analysis.
\newblock In {\em Proc. of {PLDI}}, pages 123--132, 2014.

\bibitem{Bueler2000}
B.~B{\"u}eler, A.~Enge, and K.~Fukuda.
\newblock {\em Exact Volume Computation for Polytopes: A Practical Study},
  pages 131--154.
\newblock 2000.

\bibitem{CaiLX14}
J.~Cai, P.~Lu, and M.~Xia.
\newblock The complexity of complex weighted {B}oolean {\#}{CSP}.
\newblock {\em J. Comput. Syst. Sci.}, 80(1):217--236, 2014.

\bibitem{ChakrabortyMMV16}
S.~Chakraborty, K.~S. Meel, R.~Mistry, and M.~Y. Vardi.
\newblock Approximate probabilistic inference via word-level counting.
\newblock In {\em Proc. of {AAAI}}, pages 3218--3224, 2016.

\bibitem{ChakrabortyMV13}
S.~Chakraborty, K.~S. Meel, and M.~Y. Vardi.
\newblock A scalable approximate model counter.
\newblock In {\em Proc. of {CP}}, pages 200--216, 2013.

\bibitem{ChakrabortyMV16b}
S.~Chakraborty, K.~S. Meel, and M.~Y. Vardi.
\newblock Algorithmic improvements in approximate counting for probabilistic
  inference: From linear to logarithmic {SAT} calls.
\newblock In {\em Proc. of {IJCAI}}, pages 3569--3576, 2016.

\bibitem{ChaviraD08}
M.~Chavira and A.~Darwiche.
\newblock On probabilistic inference by weighted model counting.
\newblock {\em Artificial Intelligence}, 172(6{-}7):772--799, 2008.

\bibitem{ChistikovDM15}
D.~Chistikov, R.~Dimitrova, and R.~Majumdar.
\newblock Approximate counting in {SMT} and value estimation for probabilistic
  programs.
\newblock In {\em Proc. of {TACAS}}, pages 320--334, 2015.

\bibitem{CousinsV15}
B.~Cousins and S.~S. Vempala.
\newblock Bypassing {KLS:} gaussian cooling and an $\mbox{O}^*(n^3)$ volume
  algorithm.
\newblock In {\em Proc. of {STOC}}, pages 539--548, 2015.

\bibitem{CousinsV16}
B.~Cousins and S.~S. Vempala.
\newblock A practical volume algorithm.
\newblock {\em Math. Program. Comput.}, 8(2):133--160, 2016.

\bibitem{Darwiche11}
A.~Darwiche.
\newblock {SDD:} {A} new canonical representation of propositional knowledge
  bases.
\newblock In {\em Proc. of {IJCAI}}, pages 819--826, 2011.

\bibitem{DyerF88}
M.~E. Dyer and A.~M. Frieze.
\newblock On the complexity of computing the volume of a polyhedron.
\newblock {\em {SIAM} J. Comput.}, 17(5):967--974, 1988.

\bibitem{DyerFK89}
M.~E. Dyer, A.~M. Frieze, and R.~Kannan.
\newblock A random polynomial time algorithm for approximating the volume of
  convex bodies.
\newblock In {\em Proc. of {ACM STOC}}, pages 375--381, 1989.

\bibitem{FremontRS17}
D.~J. Fremont, M.~N. Rabe, and S.~A. Seshia.
\newblock Maximum model counting.
\newblock In {\em Proc. of {AAAI}}, pages 3885--3892, 2017.

\bibitem{FukudaLM97}
K.~Fukuda, T.~M. Liebling, and F.~Margot.
\newblock Analysis of backtrack algorithms for listing all vertices and all
  faces of a convex polyhedron.
\newblock {\em Comput. Geom.}, 8:1--12, 1997.

\bibitem{GaoLZC18}
W.~Gao, H.~Lv, Q.~Zhang, and D.~Cai.
\newblock Estimating the volume of the solution space of {SMT(LIA)} constraints
  by a flat histogram method.
\newblock {\em Algorithms}, 11(9):142, 2018.

\bibitem{GeM15}
C.~Ge and F.~Ma.
\newblock A fast and practical method to estimate volumes of convex polytopes.
\newblock In {\em Proc. of {FAW}}, pages 52--65, 2015.

\bibitem{GeMLZM18}
C.~Ge, F.~Ma, T.~Liu, J.~Zhang, and X.~Ma.
\newblock A new probabilistic algorithm for approximate model counting.
\newblock In {\em Proc. of {IJCAR}}, pages 312--328, 2018.

\bibitem{GeMMZHZ19}
C.~Ge, F.~Ma, X.~Ma, F.~Zhang, P.~Huang, and J.~Zhang.
\newblock Approximating integer solution counting via space quantification for
  linear constraints.
\newblock In {\em Proc. of {IJCAI}}, pages 1697--1703, 2019.

\bibitem{GeMZZ18}
C.~Ge, F.~Ma, P.~Zhang, and J.~Zhang.
\newblock Computing and estimating the volume of the solution space of
  {SMT(LA)} constraints.
\newblock {\em TCS}, 743:110--129, 2018.

\bibitem{GeldenhuysDV12}
J.~Geldenhuys, M.~B. Dwyer, and W.~Visser.
\newblock Probabilistic symbolic execution.
\newblock In {\em Proc. of {ISSTA}}, pages 166--176, 2012.

\bibitem{GomesSS09}
C.~P. Gomes, A.~Sabharwal, and B.~Selman.
\newblock Model counting.
\newblock In {\em Handbook of Satisfiability}, pages 633--654. {IOS} Press,
  2009.

\bibitem{MI94-07}
P.~Gritzmann and V.~Klee.
\newblock On the complexity of some basic problems in computational convexity:
  {II.} volume and mixed volumes.
\newblock {\em Universit{\"{a}}t Trier, Mathematik/Informatik,
  Forschungsbericht}, 94-07, 1994.

\bibitem{HuangLOB18}
A.~Huang, L.~Lloyd, M.~Omar, and J.~C. Boerkoel.
\newblock New perspectives on flexibility in simple temporal planning.
\newblock In {\em Proc. of {ICAPS}}, pages 123--131, 2018.

\bibitem{KannanV97}
R.~Kannan and S.~Vempala.
\newblock Sampling lattice points.
\newblock In {\em {STOC}}, pages 696--700, 1997.

\bibitem{KlebanovMM13}
V.~Klebanov, N.~Manthey, and C.~Muise.
\newblock {SAT}-based analysis and quantification of information flow in
  programs.
\newblock In {\em Proc. of {QEST}}, pages 177--192, 2013.

\bibitem{LagniezLM16}
J.~Lagniez, E.~Lonca, and P.~Marquis.
\newblock Improving model counting by leveraging definability.
\newblock In {\em Proc. of {IJCAI}}, pages 751--757, 2016.

\bibitem{LiuZZ07}
S.~Liu, J.~Zhang, and B.~Zhu.
\newblock Volume computation using a direct monte carlo method.
\newblock In {\em Proc. of {COCOON}}, pages 198--209, 2007.

\bibitem{LoeraHTY04}
J.~A.~De Loera, R.~Hemmecke, J.~Tauzer, and R.~Yoshida.
\newblock Effective lattice point counting in rational convex polytopes.
\newblock {\em JSC}, 38(4):1273--1302, 2004.

\bibitem{LovaszD12}
L.~Lov{\'{a}}sz and I.~De{\'{a}}k.
\newblock Computational results of an $\mbox{O}^*(n^4)$ volume algorithm.
\newblock {\em European J. of Operational Research}, 216(1):152--161, 2012.

\bibitem{MaLZ09}
F.~Ma, S.~Liu, and J.~Zhang.
\newblock Volume computation for boolean combination of linear arithmetic
  constraints.
\newblock In {\em Proc. of {CADE-22}}, pages 453--468, 2009.

\bibitem{MuiseMBH12}
C.~J. Muise, S.~A. McIlraith, J.~Christopher Beck, and E.~I. Hsu.
\newblock Dsharp: Fast d-dnnf compilation with sharpsat.
\newblock In {\em Proc. of {Canadian AI}}, pages 356--361, 2012.

\bibitem{PesantQZ12}
G.~Pesant, C.-G. Quimper, and A.~Zanarini.
\newblock Counting-based search: Branching heuristics for constraint
  satisfaction problems.
\newblock {\em JAIR}, 43:173--210, 2012.

\bibitem{PolicellaSCO04}
N.~Policella, S.~F. Smith, A.~Cesta, and A.~Oddi.
\newblock Generating robust schedules through temporal flexibility.
\newblock In {\em Proc. of {ICAPS}}, pages 209--218, 2004.

\bibitem{Roth96}
D.~Roth.
\newblock On the hardness of approximate reasoning.
\newblock {\em Artificial Intelligence}, 82(1-2):273--302, 1996.

\bibitem{SangBK05}
T.~Sang, P.~Beame, and H.~A. Kautz.
\newblock Performing {B}ayesian inference by weighted model counting.
\newblock In {\em Proc. of {AAAI}}, pages 475--481, 2005.

\bibitem{SharmaRSM19}
S.~Sharma, S.~Roy, M.~Soos, and K.~S. Meel.
\newblock {GANAK:} {A} scalable probabilistic exact model counter.
\newblock In {\em Proc. of {IJCAI}}, pages 1169--1176, 2019.

\bibitem{SoosM19}
M.~Soos and K.~S. Meel.
\newblock {BIRD:} engineering an efficient {CNF-XOR} {SAT} solver and its
  applications to approximate model counting.
\newblock In {\em Proc. of {AAAI}}, pages 1592--1599, 2019.

\bibitem{TalvitieKNK18b}
T.~Talvitie, K.~Kangas, T.~Niinim\"{a}ki, and M.~Koivisto.
\newblock A scalable scheme for counting linear extensions.
\newblock In {\em Proc. of {IJCAI}}, pages 5119--5125, 2018.

\bibitem{TodaT15}
T.~Toda and K.~Tsuda.
\newblock {BDD} construction for all solutions {SAT} and efficient caching
  mechanism.
\newblock In {\em Proc. of {SAC}}, pages 1880--1886, 2015.

\bibitem{Valiant79}
L.~G. Valiant.
\newblock The complexity of enumeration and reliability problems.
\newblock {\em {SIAM} J. Comput.}, 8(3):410--421, 1979.

\bibitem{VerdoolaegeSBLB07}
S.~Verdoolaege, R.~Seghir, K.~Beyls, V.~Loechner, and M.~Bruynooghe.
\newblock Counting integer points in parametric polytopes using {B}arvinok's
  rational functions.
\newblock {\em Algorithmica}, 48(1):37--66, 2007.

\bibitem{ZhaoCSE16}
S.~Zhao, S.~Chaturapruek, A.~Sabharwal, and S.~Ermon.
\newblock Closing the gap between short and long xors for model counting.
\newblock In {\em Proc. of {AAAI}}, pages 3322--3329, 2016.

\bibitem{ZhouHSHCG15}
M.~Zhou, F.~He, X.~Song, S.~He, G.~Chen, and M.~Gu.
\newblock Estimating the volume of solution space for satisfiability modulo
  linear real arithmetic.
\newblock {\em Theory Comput. Syst.}, 56(2):347--371, 2015.

\end{thebibliography}

\end{document}